\documentclass{article} 
\usepackage{collas2026_conference,times}
\usepackage{easyReview}
\usepackage[normalem]{ulem}
\usepackage{booktabs}

\usepackage{amsmath,amsfonts,bm}

\def\eqref#1{equation~\ref{#1}}

\def\1{\bm{1}}

\DeclareMathAlphabet{\mathsfit}{\encodingdefault}{\sfdefault}{m}{sl}
\SetMathAlphabet{\mathsfit}{bold}{\encodingdefault}{\sfdefault}{bx}{n}

\usepackage{hyperref}
\hypersetup{
    colorlinks=true,
    linkcolor=red,
    filecolor=magenta,
    urlcolor=blue,
    citecolor=purple,
    pdftitle={Overleaf Example},
    pdfpagemode=FullScreen,
    }

\usepackage{mathtools}
\usepackage{subcaption}
\usepackage[dvipsnames]{xcolor}

\usepackage{algorithm}
\usepackage{algorithmic}

\usepackage{tikz}
\usepackage{amsmath, amssymb}
\usepackage{xcolor}

\title{Multi-Agent Empowerment \\ and Emergence of Complex Behavior in Groups}

\author{%
  \begin{minipage}[t]{0.40\textwidth}
    \normalfont\raggedright
    \textbf{Tristan Shah} \\
    Department of Computer Science \\
    Texas Tech University \\
    United States of America \\
    \texttt{trisshah@ttu.edu}
  \end{minipage}%
  \hspace{0.15\textwidth}%
  \begin{minipage}[t]{0.45\textwidth}
    \normalfont\raggedright
    \textbf{Ilya Nemenman} \\
    Department of Physics, Department of Biology, \\
    Initiative in Theory and Modeling of Living Systems \\
    Emory University \\
    United States of America \\
    \texttt{ilya.nemenman@emory.edu}
  \end{minipage}%
  \vspace{2em}%
  \\
  \begin{minipage}[t]{0.40\textwidth}
    \normalfont\raggedright
    \textbf{Daniel Polani} \\
    Department of Computer Science, School of Physics, Engineering and Computer Science
    University of Hertfordshire \\
    United Kingdom \\
    \texttt{d.polani@herts.ac.uk}
  \end{minipage}%
  \hspace{0.15\textwidth}%
  \begin{minipage}[t]{0.45\textwidth}
    \normalfont\raggedright
    \textbf{Stas Tiomkin}\thanks{Corresponding author} \\
    Department of Computer Science \\
    Texas Tech University \\
    United States of America \\
    \texttt{stas.tiomkin@ttu.edu}
  \end{minipage}%
}

\collasfinalcopy 

\begin{document}

\maketitle

\begin{abstract}
  Intrinsic motivations are receiving increasing attention, i.e. behavioral incentives that are not engineered, but emerge from the interaction of an agent with its surroundings. In this work we study the emergence of behaviors driven by one such incentive, empowerment, specifically in the context of more than one agent. We formulate a principled extension of empowerment to the multi-agent setting, and demonstrate its efficient calculation. We observe that this intrinsic motivation gives rise to characteristic modes of group-organization in two qualitatively distinct environments: a pair of agents coupled by a tendon, and a controllable Vicsek flock. This demonstrates the potential of intrinsic motivations such as empowerment to not just drive behavior for only individual agents but also higher levels of behavioral organization at scale. \\ \\ A project website and code is available at \href{https://fanciful-palmier-2afc73.netlify.app/}{this anonymous link}.
\end{abstract}

\section{Introduction}

Explicit supervision through engineered rewards \citep{sutton1998reinforcement} and expert demonstrations \citep{hussein2017imitation} remain the standard framework for training artificial agents. These signals are specific to particular tasks, hard to design canonically, and yield brittle task-specific solutions \citep{amodei2016concrete, krakovna2020avoiding}. Biological organisms, in contrast, rarely have well-delineated extrinsic tasks, yet they continuously learn and adapt \citep{oudeyer2007intrinsic}. A growing body of work instead explores \emph{intrinsic motivations} \citep{gregor2016variational, tiomkin_2024, zheng2025can} --- principles that generate behavior from an agent's interaction with its surroundings, without reference to external tasks. Intrinsically motivated behaviors are generic: they emerge from the structure of the agent-environment coupling rather than from a designer's choice of task. Such generic behaviors can then be combined and refined into more complex task-specific behaviors \citep{sharma2020dynamicsawareunsuperviseddiscoveryskills, laskin2021urlb, eysenbach2018diversity}; they can also equip an agent with a local incentive structure that cushions it against random or adversarial disturbances that would derail a naively task-optimized strategy \citep{tiomkin2025process}. This makes intrinsic motivation a natural substrate for asking what behaviors interacting agents produce on their own, without shared objectives --- the question we study here.


As autonomous agents become increasingly prevalent in virtual and physical environments, understanding what behaviors can emerge in agent groups from purely intrinsic motivation becomes an important question. We ask: when agents pursue only their own intrinsic motivation, with no explicit coordination or shared objective, what behaviors, if any, emerge from their interaction, and can we understand why? And how do those behaviors change when an agent instead directs that same capacity toward helping another?

We model intrinsic motivation as maximization of empowerment \citep{klyubin2005empowerment, salge2014empowerment}, defined as the mutual information between an agent's sequence of actions and the subsequent future observations of the world. Informally, a highly empowered agent has the potential to have a large effect on the world. An egoistic agent then chooses actions that maximize its own empowerment. An altruistic agent instead chooses actions that maximize the empowerment of another agent. This makes empowerment a natural framework for studying how the direction of intrinsic motivation shapes group behavior.

Prior work has mostly studied empowerment in single-agent settings, leaving its behavior under multi-agent interaction largely unexplored (though see \citealt{capdepuy2012perception, capdepuy2007maximization, salge2017empowerment, charlesworth_flocking}). Recently, \citet{tiomkin_2024} proposed a method for calculating empowerment for arbitrary durations of action sequences and future world sequences using a weak-control approximation around the system's autonomous dynamics, and mapped the problem onto Shannon's classical Gaussian channel. Here we extend these results to the multi-agent case by modeling coupled dynamics as a multi-user interference channel \citep{carleial_1978}. In our proposed approach (Figure~\ref{fig:interference_channel}), each agent treats the actions of others as structured interference noise and solves for its optimal strategy via iterative water-filling \citep{yu_2004}, reaching a Nash equilibrium. This yields a tractable approximation of empowerment in multi-agent systems without requiring explicit coordination or communication between agents.

We study this framework in two qualitatively different classes of systems. The first is the simplest case: a pair of mechanically coupled agents, where empowerment maximization produces dominance hierarchies when agents have unequal strength and cooperation when their strengths are comparable. At the other extreme, we study a large group of agents, modeled as a Vicsek-style flock \citep{vicsek_1995}. Here, egoistic empowerment prevents the usual convergence to a shared heading and instead drives the population into two opposing directional bands. Together these results suggest that empowerment maximization, without any explicit social objective, can generate non-trivial group-level behaviors whose character depends sensitively on agents' relative capabilities and the nature of their coupling.

Our main contributions are as follows: (1) formulation of multi-agent empowerment as an information-theoretic interference channel, (2) efficient solution of the resulting optimization via iterative water-filling, and (3) demonstration of the diversity of behaviors emerging via multi-agent empowerment maximization in few- and many-agent systems.

\section{Related Work}

\subsection{Intrinsically Motivated Collective Motion}

Collective motion in animal groups has long served as a testbed for
studying how simple local interactions give rise to group-level
structure
\citep{couzin2009collective,cavagna2014bird,ouellette2022physics}.
\cite{reynolds_1987} showed that rich flocking dynamics can emerge
from a few local decision rules, and \cite{vicsek_1995} distilled
this idea into a minimal active-matter model that has since become a
standard benchmark. Informational accounts of such rules, deriving
them from local information maximization rather than prescribing them
by hand, have also been proposed
\citep{salge11:_local_infor_maxim_emerg_flock_behav}.

\cite{charlesworth_flocking} studied collective behavior under Future State Maximization (FSM), where each agent acts to maximize the number of distinct visual futures it can reach over a finite time horizon. Agents perceive neighbors through a simple retina and choose among a small set of discrete actions by explicitly enumerating reachable future percepts along a rollout tree. Cohesion, co-alignment, and collision avoidance, which all are qualitative hallmarks of animal flocking, emerge spontaneously, without being built into the objective. Because explicit enumeration scales poorly, the authors trained a neural network to approximate FSM and reach larger flock sizes.

The principle of FSM overlaps substantially with empowerment
maximization: both objectives reward agents for preserving a large
set of reachable futures. However, the two frameworks diverge in
important respects. First, our method is analytically tractable: by
casting multi-agent empowerment as an interference channel and
solving via iterative water-filling, we obtain a principled solution
without learned approximations. Second, our agents operate directly
in continuous action and state spaces with coupled dynamics, rather
than over discrete visual percepts and a fixed search tree. We will
furthermore find that the precise choice of ``sensors'' --- i.e.,
what the agent observes about its reachable futures --- matters. In
the experiments presented here, the collective behaviors emerging
under FSM and under empowerment differ markedly: FSM drives
consensus alignment and cohesion, whereas egoistic empowerment
produces a bimodal configuration in which agents self-organize into
opposing directional bands.

\subsection{Social Influence as Intrinsic Motivation}

\cite{jaques2019socialinfluenceintrinsicmotivation} introduce social
influence as an intrinsic reward: each agent is rewarded for
maximizing its causal effect on other agents' actions, regardless of
whether this benefits itself or its targets. The reward is estimated
via a counterfactual over the other agents' action distributions and
used to train neural-network policies with deep reinforcement
learning. Our framework differs from this one in the target of the
objective. Each of our agents seeks to maximize control over its own
future states, treating others as structured interference rather than
as targets of influence. Egoistic and altruistic behaviors then arise
as natural choices of whose empowerment to maximize --- a degree of
freedom absent in the social-influence framework. Empowerment could
in principle also quantify influence over others, which would bring
the two frameworks into closer alignment; we leave that direction for
future work. As in the comparison with FSM above, our formulation
remains analytically tractable and avoids learned approximations.

\section{Background}\label{sec:single-agent}


For a single agent, empowerment \citep{klyubin2005empowerment,
salge2014empowerment} measures how much influence an agent has over
its future: formally, it is the channel capacity between an agent's
open-loop action sequence $U_{0:t-1} \in \mathbb{R}^{t \cdot d_u}$
and the resulting future state $X_t \in \mathbb{R}^{d_x}$ (one can
also generalize to state sequences \citep{tiomkin_2024}).
In general, an agent need not care about its entire future state. A
\emph{sensor} $C \in \mathbb{R}^{d_y \times d_x}$ selects the
components the agent seeks to control, producing the sensed output
$Y_t = CX_t \in \mathbb{R}^{d_y}$. Empowerment is then the channel
capacity between the action sequence and the sensed future state,
maximized over the agent's probing policy $\pi(\cdot | x_0)$:
\begin{equation*}
    \mathcal{E}(x_0) \coloneq \max_{\pi(\cdot |x_0)}
    \mathcal{I}[Y_t; U_{0:t-1} \mid X_0 = x_0].
\end{equation*}
A tractable approximation proceeds in four steps: (1) treat the
action sequence as a small perturbation to the nonlinear autonomous
dynamics and retain only the linear response, (2) assume the probing
policy is Gaussian,
\begin{equation*}
    \pi(U_{0:t-1} | X_0 = x_0) \coloneq \mathcal{N}(\mathbf{0},
    \Sigma_{u_{0:t-1}}),
\end{equation*}
(3) add observation noise $z \sim \mathcal{N}(\mathbf{0}, \Sigma_z)$
to model a realistic noisy sensor (ensuring finite channel capacity),
and (4) apply the water-filling algorithm for a power-constrained
linear Gaussian channel \citep{cover1999elements}. This gives:
\begin{equation*}
    \mathcal{E}(x_0) \approx \max_{\Sigma_{u_{0:t-1}} \succeq 0} \frac{1}{2}
    \ln \big| CF_t \Sigma_{u_{0:t-1}} F_t^\top C^\top +
    \Sigma_z \big| - \frac{1}{2} \ln \big| \Sigma_z \big|, \quad \textrm{s.t.} \quad
    \mathrm{tr}(\Sigma_{u_{0:t-1}}) \leq P,
\end{equation*}
where $F_t$ is the sensitivity of the future state to the control
sequence along the autonomous trajectory,
\begin{equation*}
    F_t(x_0) \coloneq \frac{\partial x_t}{\partial
    u_{0:t-1}}\bigg|_{u_{0:t-1} = \mathbf{0}} \in \mathbb{R}^{d_x
    \times (t \cdot d_u)},
\end{equation*}
so that $CF_t \in \mathbb{R}^{d_y \times (t \cdot d_u)}$ is the
effective channel from actions to sensed output. Water-filling
distributes power across the eigenmodes of this effective channel.
When $C = \mathbf{I}$ the sensor is the full state and the original
formulation is recovered. This single-agent formulation does not
extend directly to multiple interacting agents; we address this in Section \ref{sec:method}.

\section{Method}\label{sec:method}

\begin{figure}
    \centering
    \includegraphics[width=\linewidth]{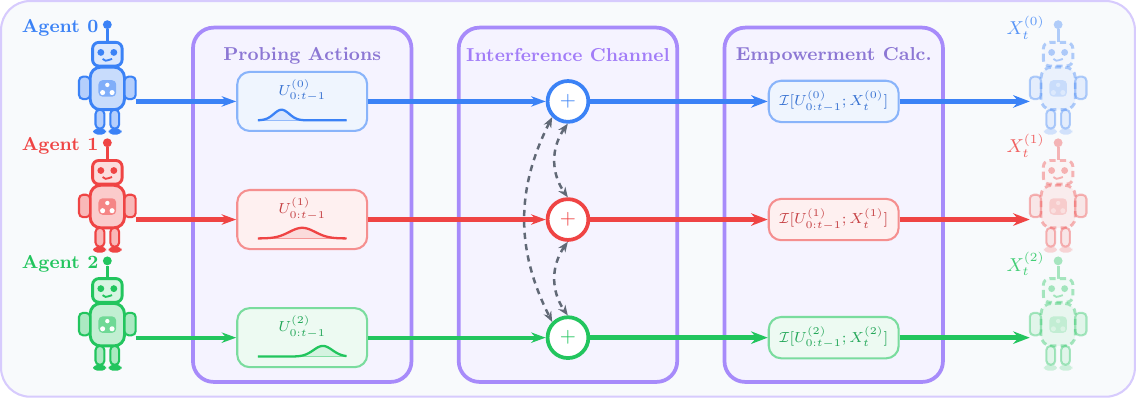}
    \caption{Multi-agent system as an interference channel, illustrated for $N=3$ agents with identity sensors $C^{(n)} = \mathbf{I}$. Each agent $n$ transmits its action sequence $U_{0:t-1}^{(n)}$ to its own future state $X_t^{(n)}$, which serves as its receiver; the other agents' actions appear as structured interference, reducing each agent's capacity to influence its future state.}
    \label{fig:interference_channel}
\end{figure}

In this work we extend empowerment to multi-agent systems with many
interacting agents (Figure \ref{fig:interference_channel}). A key challenge is that channel capacity, the
quantity underlying empowerment, is generally intractable for
nonlinear dynamics. We propose a multi-agent extension of
linear-Gaussian empowerment by decomposing the coupled linearized
dynamics around the autonomous nonlinear solution into an
interference-channel structure: each agent treats its own actions as
a signal to its own future state, and the other agents' actions as
noise. This formulation inherits standard solution techniques and
guarantees from multi-user information theory, extending the
single-agent construction of Section~\ref{sec:single-agent} to the
coupled case. Each agent's empowerment is then the capacity of its
own single-user channel under this structured interference. The
agents' probing distributions are determined jointly as the Nash
equilibrium of a non-cooperative game, which we compute by (1)
linearizing the coupled dynamics over a finite planning horizon, (2)
solving for the per-agent probing covariances with the iterative
water-filling (IWF) algorithm \citep{yu_2004}, and (3) selecting
per-agent control actions by empowerment maximization.

\subsection{Multi-Agent Linearization}\label{sec:multi-linearization}

We consider a multi-agent system composed of $N$ agents indexed by $n \in \{0, \ldots, N-1\}$. Let $x_0^{(n)}$ denote the state of agent $n$ at time $0$, and $u_0^{(n)}$ its action. We stack the individual components of all agents into joint column vectors:
\begin{equation*}
    \mathbf{x}_0 =
    \begin{bmatrix}
        x_0^{(0)} \\ x_0^{(1)} \\ \vdots \\ x_0^{(N-1)}
    \end{bmatrix}, \quad
    \mathbf{u}_0 =
    \begin{bmatrix}
        u_0^{(0)} \\ u_0^{(1)} \\ \vdots \\ u_0^{(N-1)}
    \end{bmatrix}, \quad
    \mathbf{u}_{0:t-1} =
    \begin{bmatrix}
        u_{0:t-1}^{(0)} \\ u_{0:t-1}^{(1)} \\ \vdots \\ u_{0:t-1}^{(N-1)}
    \end{bmatrix},
\end{equation*}
where $u_{0:t-1}^{(n)}$ denotes the action sequence of agent $n$ over the planning horizon $0$ to $t-1$, and $\mathbf{u}_{0:t-1}$ is the concatenation of all agents' sequences. The system evolves in discrete time under coupled, generally nonlinear dynamics:
\begin{equation*}
    \mathbf{x}_{k+1} = f(\mathbf{x}_k, \mathbf{u}_k).
\end{equation*}
To reduce the multi-agent empowerment problem to a tractable linear-Gaussian form, we linearize these dynamics around the autonomous (zero-control) trajectory. The sensitivity of the joint state at horizon $t$ to the joint action sequence is the block Jacobian
\begin{equation}\label{eq:block_sensitivity}
    \mathbf{F}_t \coloneq
    \begin{bmatrix}
        F_t^{(0, 0)} & F_t^{(0, 1)} & \cdots & F_t^{(0, N-1)} \\
        F_t^{(1, 0)} & F_t^{(1, 1)} & \cdots & F_t^{(1, N-1)} \\
        \vdots & \vdots & \ddots & \vdots \\
        F_t^{(N-1, 0)} & F_t^{(N-1, 1)} & \cdots & F_t^{(N-1, N-1)}
    \end{bmatrix}, \quad F_t^{(n, m)} \coloneq \frac{\partial x_t^{(n)}}{\partial u_{0:t-1}^{(m)}}\bigg|_{\mathbf{u}_{0:t-1}=\mathbf{0}} \in \mathbb{R}^{d_x \times (t \cdot d_u)}.
\end{equation}
The diagonal blocks $F_t^{(n,n)}$ describe each agent's direct influence on its own future state; the off-diagonal blocks $F_t^{(n,m)}$ with $m \neq n$ capture the influence of agent $m$ on agent $n$'s future state through the coupling. This block structure is the basis for the interference-channel formulation developed next.

Each agent $n$ is equipped with a sensor $C^{(n)} \in \mathbb{R}^{d_y \times d_x}$ that picks out the aspects of its future state the agent seeks to influence, yielding the sensed output $Y_t^{(n)} = C^{(n)} X_t^{(n)} \in \mathbb{R}^{d_y}$. The effective channel from agent $n$'s own actions to its sensed output is then $C^{(n)} F_t^{(n,n)}$, while the interference arriving from any other agent $m \neq n$ passes through $C^{(n)} F_t^{(n,m)}$. These are the multi-agent analogs of the effective channel $CF_t$ from Section~\ref{sec:single-agent}, now split by source agent into a direct (signal) block and off-diagonal interference blocks.

\subsection{Iterative Water-Filling}

The linearized multi-agent system induces an interference channel \citep{carleial_1978} (Figure~\ref{fig:interference_channel}): each agent $n$ transmits through its control sequence to its own sensed output $Y_t^{(n)}$, and because the dynamics are coupled, the controls of any other agent $m \neq n$ also reach $Y_t^{(n)}$ as structured interference through the off-diagonal blocks of $\mathbf{F}_t$. This is in contrast to the multiple-access setting studied by \citet{yu_2004}, in which several transmitters share a single receiver. In our framework each agent seeks to control only its own future, each has its own receiver, so the interference-channel formulation is the natural one.


The resulting optimization is a non-cooperative game whose Nash equilibrium
characterizes the optimal probing distributions: a Nash equilibrium is a joint
probing distribution $\{S_t^{(n)}\}_{n=0}^{N-1}$ in which each agent's
covariance is a best response to the others', so that no agent can
unilaterally increase its own empowerment by changing its probing
distribution. We solve for this equilibrium with iterative water-filling
(IWF) \citep{yu_2004} (Algorithm~\ref{alg:iwf_n_agent}). At each inner
iteration, agent $n$ treats all other agents' current probing covariances
as fixed, assembles its effective noise covariance $\Sigma_{z,t}^{(n)}$
(\eqref{eq:noise-cov}), and solves the resulting single-user problem
(\eqref{eq:agent-cov-update}). The solution is precisely agent $n$'s best response
to the others' current strategies, making IWF a realization of best-response dynamics, which, under the sufficient conditions of \protect\cite{wang2025convergenceiterativewaterfillingmultiuser}, converges to a unique fixed point from any initialization.

For each agent $n$, write $F_t^{(n)} \coloneq F_t^{(n,n)}$ for the direct-channel block, and let
\begin{equation}\label{eq:noise-cov}
    \Sigma_{z,t}^{(n)} \coloneq S_z^{(n)} + \sum_{m \neq n} C^{(n)} F_t^{(n,m)} S_t^{(m)} F_t^{(n,m)\top} C^{(n)\top}
\end{equation}
be the effective noise covariance seen by agent $n$ in its sensed-output space, combining the fixed observation noise $S_z^{(n)} \in \mathbb{R}^{d_y \times d_y}$ with the interference from the other agents' probing covariances $\{S_t^{(m)}\}_{m \neq n}$. Each agent's probing covariance is subject to the power budget $\mathrm{tr}(S_t^{(n)}) \leq P^{(n)}$. At each iteration, agent $n$ updates its action covariance by solving the single-user Gaussian channel problem
\begin{equation}\label{eq:agent-cov-update}
    S_t^{(n)} = \arg\max_{S \succeq 0}
    \frac{1}{2} \ln \left| C^{(n)} F_t^{(n)} S F_t^{(n)\top} C^{(n)\top} + \Sigma_{z,t}^{(n)} \right|,
    \quad \mathrm{s.t.} \quad \mathrm{tr}(S) \leq P^{(n)},
\end{equation}
whose solution is given by the standard water-filling algorithm \citep{cover1999elements}. After convergence, agent $n$'s empowerment is
\begin{equation}\label{eq:agent-empowerment-calculation}
    \mathcal{E}_t^{(n)} = \frac{1}{2} \ln \left| C^{(n)} F_t^{(n)} S_t^{(n)} F_t^{(n)\top} C^{(n)\top} + \Sigma_{z,t}^{(n)} \right| - \frac{1}{2} \ln \left| \Sigma_{z,t}^{(n)} \right|.
\end{equation}
At the Nash equilibrium, each agent's probing distribution is a best response to the others', maximizing its empowerment $\mathcal{E}_t^{(n)}$ given the other agents' distributions.

\begin{algorithm}
\caption{Iterative Water-Filling for $N$ Agents}
\label{alg:iwf_n_agent}
\begin{algorithmic}[1]
\REQUIRE Sensitivity matrix $\mathbf{F}_t$, power constraints $\{P^{(n)}\}_{n=0}^{N-1}$, observation noise covariances $\{S_z^{(n)}\}_{n=0}^{N-1}$, \\ and sensors $\{C^{(n)}\}_{n=0}^{N-1}$

\STATE Initialize $\{S_t^{(n)}\}_{n=0}^{N-1}$ with $\mathrm{tr}(S_t^{(n)}) \leq P^{(n)}$
\REPEAT
    \FOR{$n = 0,\dots,N-1$}
        \STATE $S_t^{(n),\mathrm{prev}} \leftarrow S_t^{(n)}$
        \STATE Compute $\Sigma_{z,t}^{(n)}$ (\eqref{eq:noise-cov})
        \STATE Update $S_t^{(n)}$ (\eqref{eq:agent-cov-update})
    \ENDFOR
\UNTIL{$\|S_t^{(n)} - S_t^{(n),\mathrm{prev}}\| \leq \epsilon$ for all $n$}
\STATE Evaluate $\{\mathcal{E}_t^{(n)}\}_{n=0}^{N-1}$ (\eqref{eq:agent-empowerment-calculation})
\RETURN $\{\mathcal{E}_t^{(n)}\}_{n=0}^{N-1}$
\end{algorithmic}
\end{algorithm}

The dominant cost per IWF iteration is assembling $\Sigma_{z,t}^{(n)}$
(\eqref{eq:noise-cov}) for each agent, which requires summing $N-1$
matrix products; across all $N$ agents this gives $\mathcal{O}(N^2)$ cost
per iteration. The per-agent water-filling step (\eqref{eq:agent-cov-update})
requires an SVD of the effective channel $C^{(n)}F_t^{(n)}$, whose cost
scales with the planning horizon $t$ and control dimension $d_u$ and state dimension $d_x$ but is
independent of $N$.

\subsection{Policy for Control}

Once the empowerment of each agent has been computed via iterative water-filling, actions can be derived for each agent. We consider two alternative control strategies:
\begin{enumerate}
    \item \textbf{Egoistic}: an agent maximizes its own empowerment.
    \item \textbf{Altruistic}: an agent maximizes the empowerment of another agent $m \neq n$.
\end{enumerate}

Under the egoistic policy, each agent selects the action that maximizes its own empowerment at the next state, with all other agents' actions held fixed at zero. Under the altruistic policy, agent $n$ instead selects the action that maximizes the empowerment of an explicitly specified agent $m$. Notably, this does not maximize agent $n$'s ability to influence agent $m$'s state; it maximizes agent $m$'s empowerment directly. As in the egoistic case, all other agents' actions are assumed to be zero during the maximization. The two policies for egoistic and altruistic empowerment-maximizing behavior are given by \eqref{eq:egoistic-policy} and \eqref{eq:altruistic-policy} respectively.
\begin{align}
    \pi_\textrm{egoistic}^{(n)}(\mathbf{x}_{k})   &= \arg\max_{u_{k}^{(n)}} \mathcal{E}_t^{(n)}\big(f(\mathbf{x}_{k}, \mathbf{u}_{k})\big) \label{eq:egoistic-policy}\\
    \pi_\textrm{altruistic}^{(n)}(\mathbf{x}_{k}) &= \arg\max_{u_{k}^{(n)}} \mathcal{E}_t^{(m)}\big(f(\mathbf{x}_{k}, \mathbf{u}_{k})\big) \label{eq:altruistic-policy}
\end{align}
where $k$ is the simulation step and $t$ is the empowerment planning horizon. Dependency on agent $n$'s action enters through the stacked joint vector $\mathbf{u}_k$ defined in Section~\ref{sec:multi-linearization}; agent $n$ optimizes its own component with others held at zero. The dynamics $f(\mathbf{x}_k, \mathbf{u}_k)$ determines the empowerment (\eqref{eq:agent-empowerment-calculation}) in the next state. This is exactly the formulation employed in previous works \citep{tiomkin_2024, salge2014empowerment}.

The standard procedure for empowerment-maximizing agents consists of two phases: (1) evaluation of empowerment (2) deriving an action to maximize empowerment. In practice (2) is achieved by selecting actions which follow the gradient of empowerment evaluated in the current state \citep{tiomkin_2024, salge2014empowerment}. This strategy is a simple and practical realization of \eqref{eq:egoistic-policy} and \eqref{eq:altruistic-policy}.

The complete online control procedure---linearization, empowerment evaluation via Algorithm~\ref{alg:iwf_n_agent}, policy selection, and dynamics step---is summarized in Algorithm~\ref{alg:interaction}.

\begin{algorithm}[h]
\caption{Online Multi-Agent Iterative Water-Filling for Control}
\label{alg:interaction}
\begin{algorithmic}[1]
\REQUIRE Initial joint state $\mathbf{x}_0$, planning horizon $t$, sensor noise covariance $\{S_z^{(n)}\}_{n=0}^{N-1}$, sensors $\{C^{(n)}\}_{n=0}^{N-1}$, and episode length T

\FOR{$k = 0$ to $T-1$}
    \STATE Linearize the $t$-step dynamics at $\mathbf{x}_k$ and compute $\mathbf{F}_t$
    \STATE Calculate each agent's empowerment $\{\mathcal{E}_t^{(n)}\}_{n=0}^{N-1}$ by algorithm \ref{alg:iwf_n_agent}
    \STATE Select action for each agent $\mathbf{u}_k$ using either \eqref{eq:egoistic-policy} or \eqref{eq:altruistic-policy}
    \STATE $\mathbf{x}_{k+1} \leftarrow f(\mathbf{x}_k, \mathbf{u}_k)$
\ENDFOR
\end{algorithmic}
\end{algorithm}

\section{Results}

We evaluate whether multi-agent empowerment induces collective behavior in coupled systems, using two distinct environments: a small-scale ($N=2$) mechanically coupled pendulum system and a large-scale ($N=125$) Vicsek-style flock with spatially local interactions. In the pendulum system, agents are tightly coupled, so one agent's influence on another is immediate and explicit. The flocking setting is more subtle: agents interact only with their local neighbors, yet non-trivial global structure emerges across agents that are never in direct contact, as information propagates through the network of transient local connections. Across both domains, our method generates interpretable individual and group-level structure under different control policies.

\subsection{Linked Pendulums}\label{sec:pendulum}

\begin{figure}
    \centering

    \begin{subfigure}[t]{0.24\columnwidth}
        \centering
        \includegraphics[width=\linewidth]{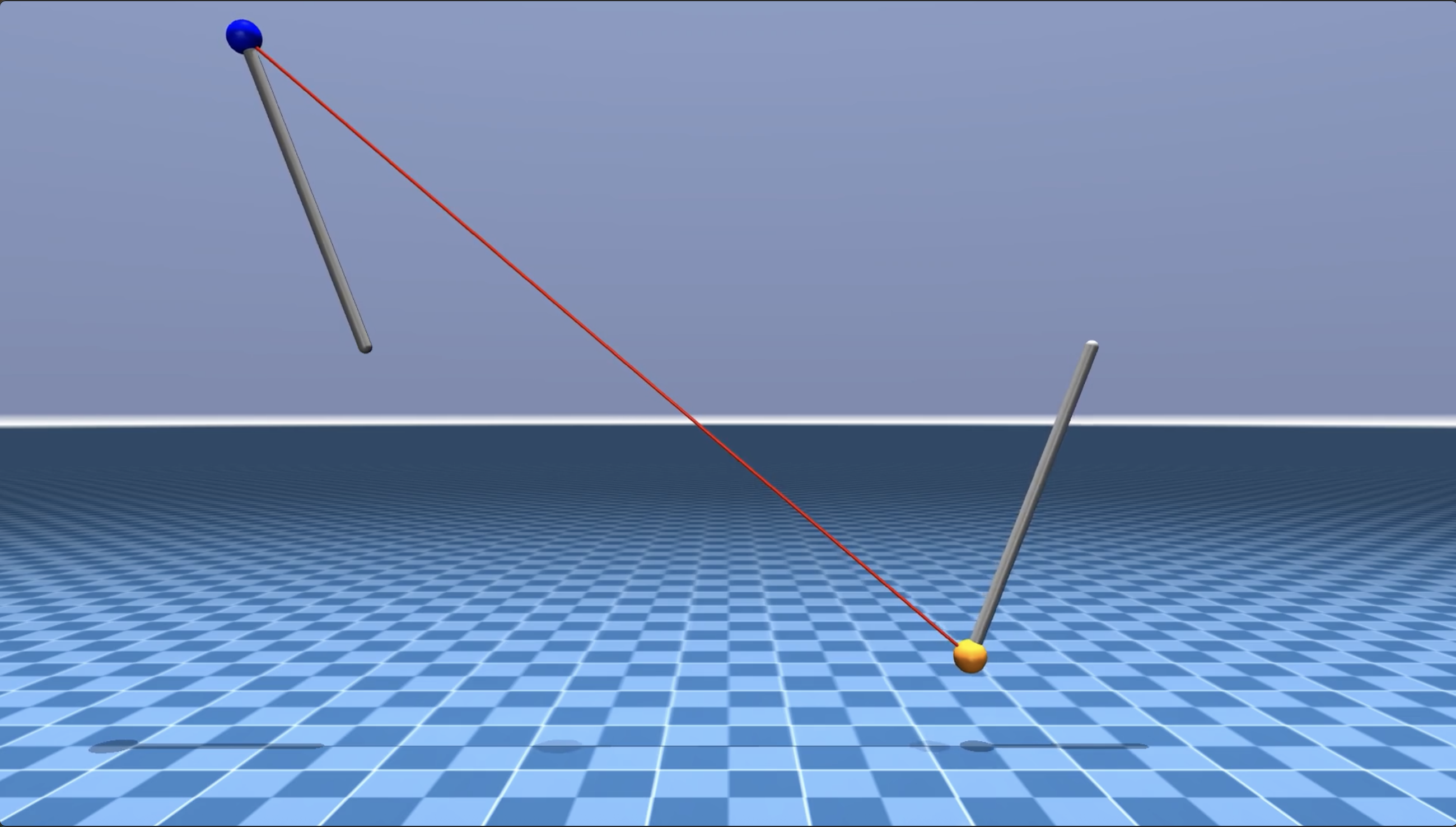}
        \caption{{\color{blue} Left Up}}
        \label{fig:left-up}
    \end{subfigure}
    \hfill
    \begin{subfigure}[t]{0.24\columnwidth}
        \centering
        \includegraphics[width=\linewidth]{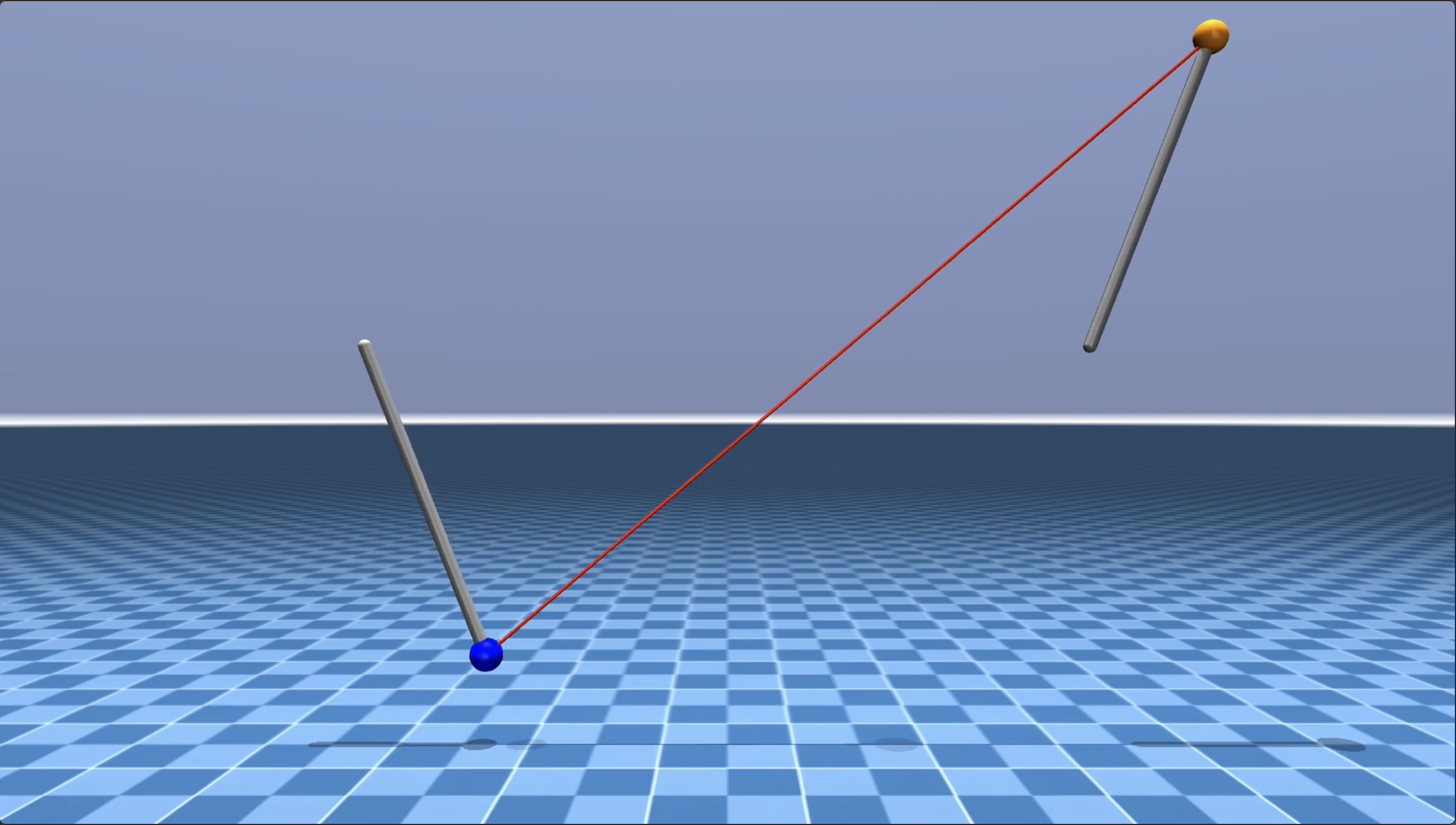}
        \caption{{\color{orange} Right Up}}
        \label{fig:right-up}
    \end{subfigure}
    \hfill
    \begin{subfigure}[t]{0.24\columnwidth}
        \centering
        \includegraphics[width=\linewidth]{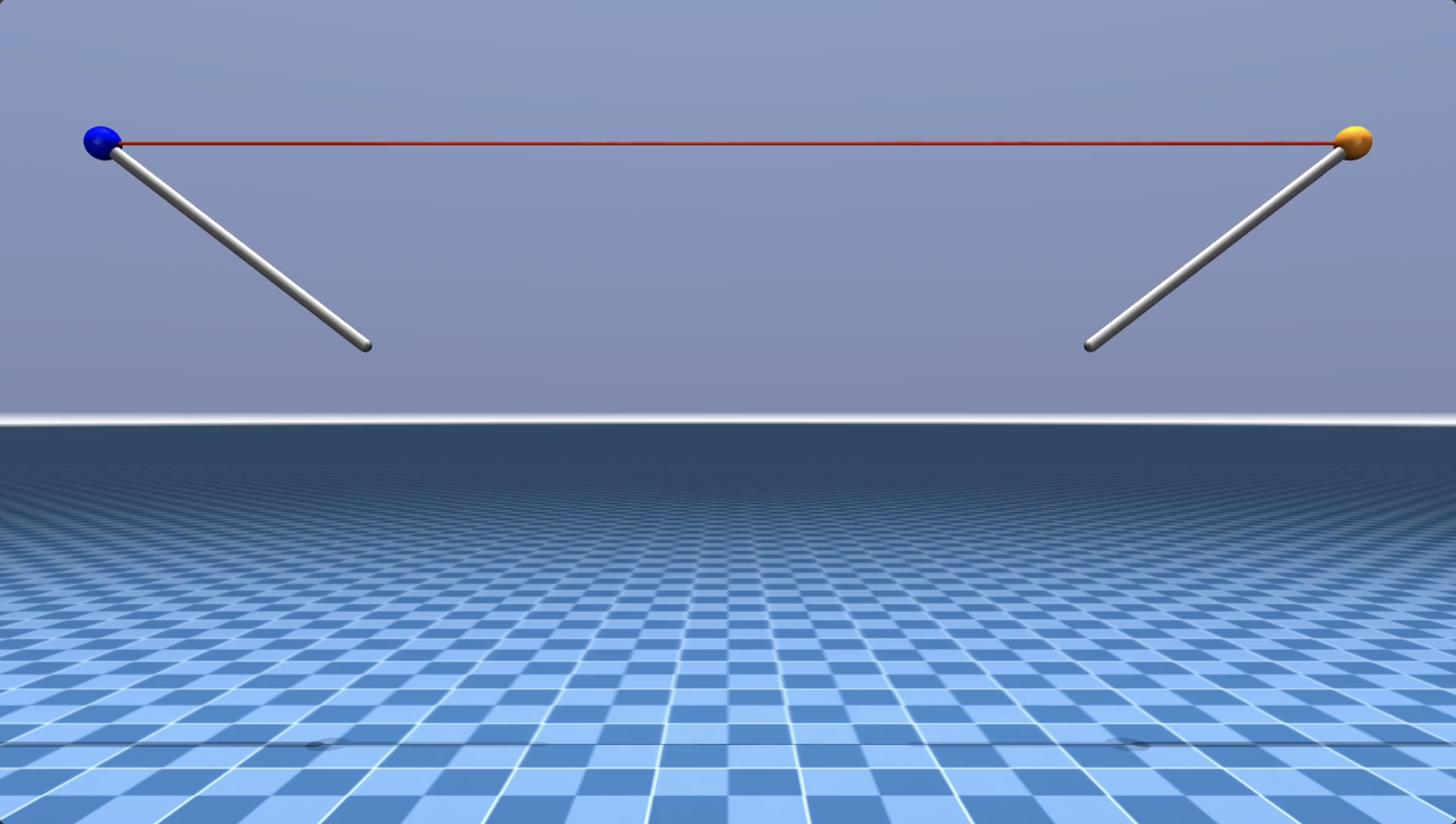}
        \caption{{\color{OliveGreen} Both Up}}
        \label{fig:both-up}
    \end{subfigure}
    \hfill
    \begin{subfigure}[t]{0.24\columnwidth}
        \centering
        \includegraphics[width=\linewidth]{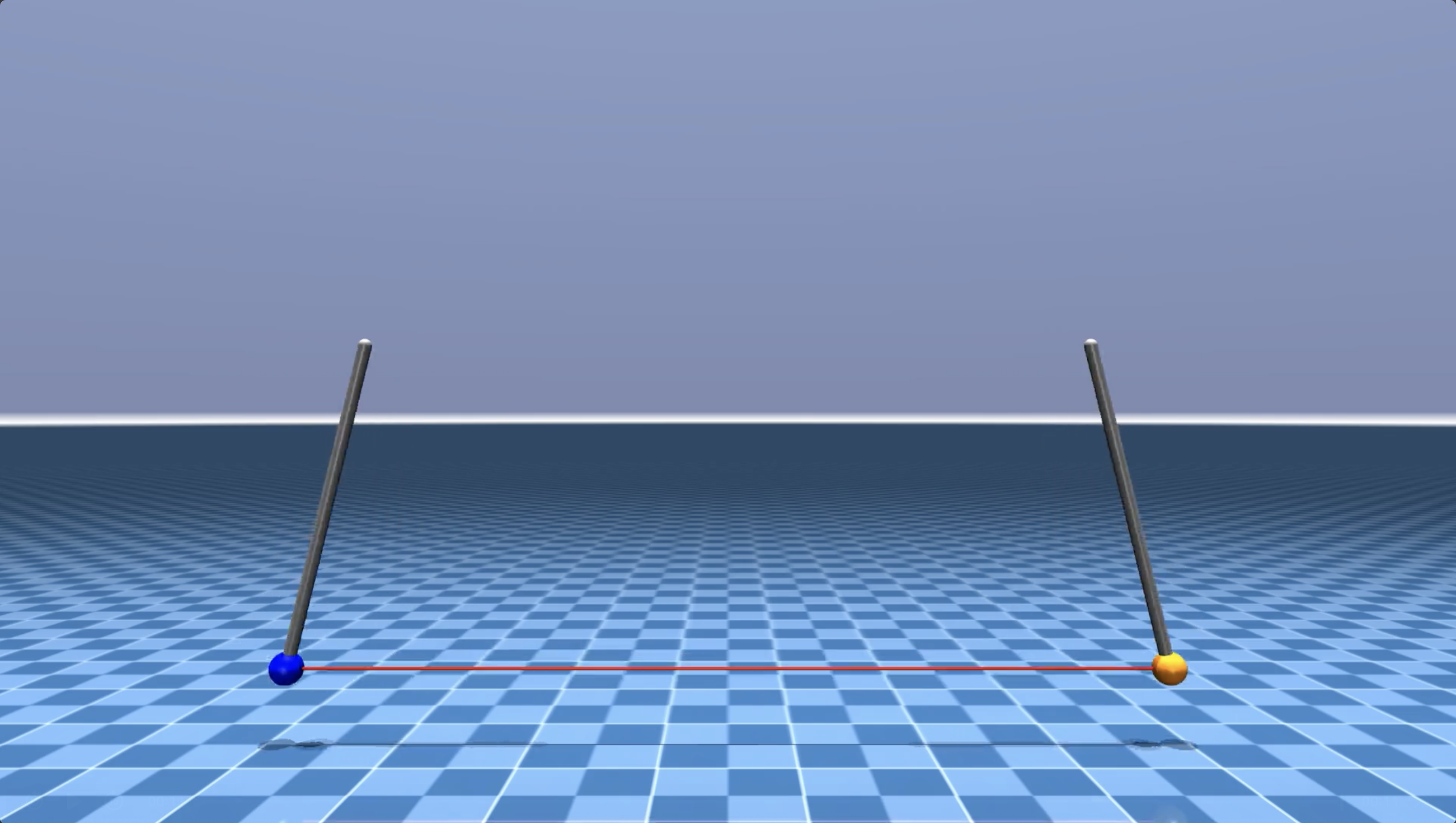}
        \caption{{\color{gray} Neither Up}}
        \label{fig:neither-up}
    \end{subfigure}

    \caption{Four outcomes observed in the linked pendulum environment. In each panel, agent $n=0$ is blue (left) and agent $n=1$ is orange (right). In Figures~\ref{fig:left-up} and~\ref{fig:right-up}, one strong agent dominates. In Figure~\ref{fig:both-up}, comparable-strength agents reach upright together. In Figure~\ref{fig:neither-up}, neither agent reaches upright.}
    \label{fig:four_side_by_side}
\end{figure}

We first study a two-agent linked-pendulum system in which each agent applies torque only to its own hinge, while the pendulum tips are coupled by an elastic tendon (shown by the red bar in Figure \ref{fig:four_side_by_side}). Each agent's sensor $C^{(n)}$ selects its angle, so empowerment reflects each agent's capacity to influence its own angle. This setting provides a minimal testbed for how empowerment depends on the agents' relative actuation strength.

\begin{figure}[t]
    \centering

    \begin{subfigure}[t]{\columnwidth}
        \centering
        \includegraphics[width=\linewidth]{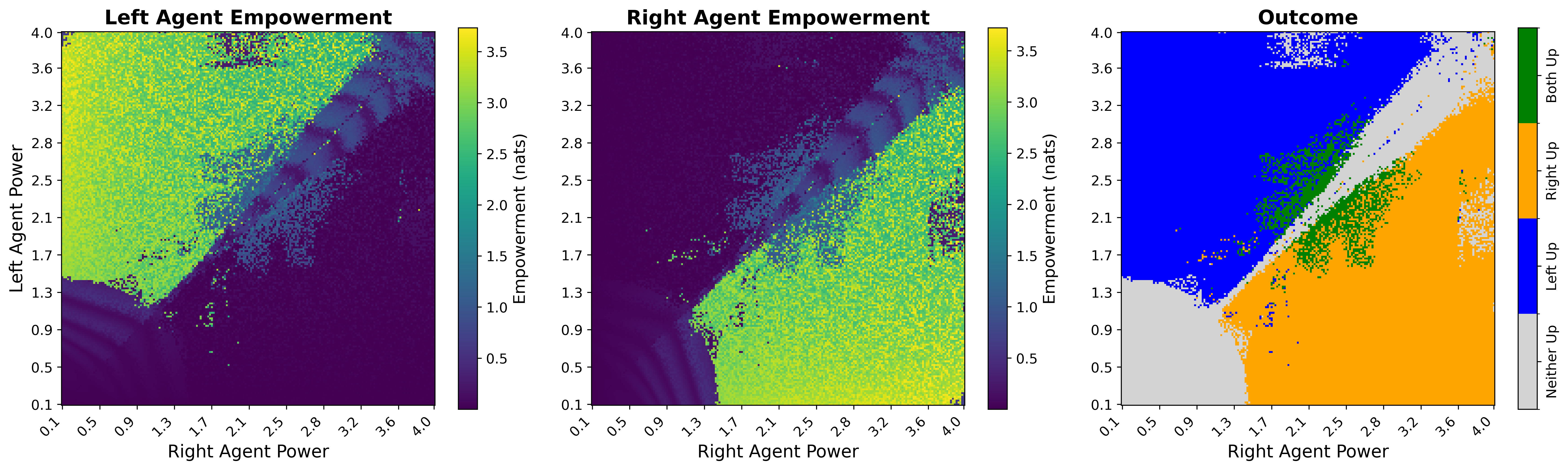}
        \caption{\textbf{Egoistic}: each agent maximizes its own empowerment. The policy is given by \eqref{eq:egoistic-policy}.}
        \label{fig:heatmap_egoistic}
    \end{subfigure}

    \vspace{0.5em}

    \begin{subfigure}[t]{\columnwidth}
        \centering
        \includegraphics[width=\linewidth]{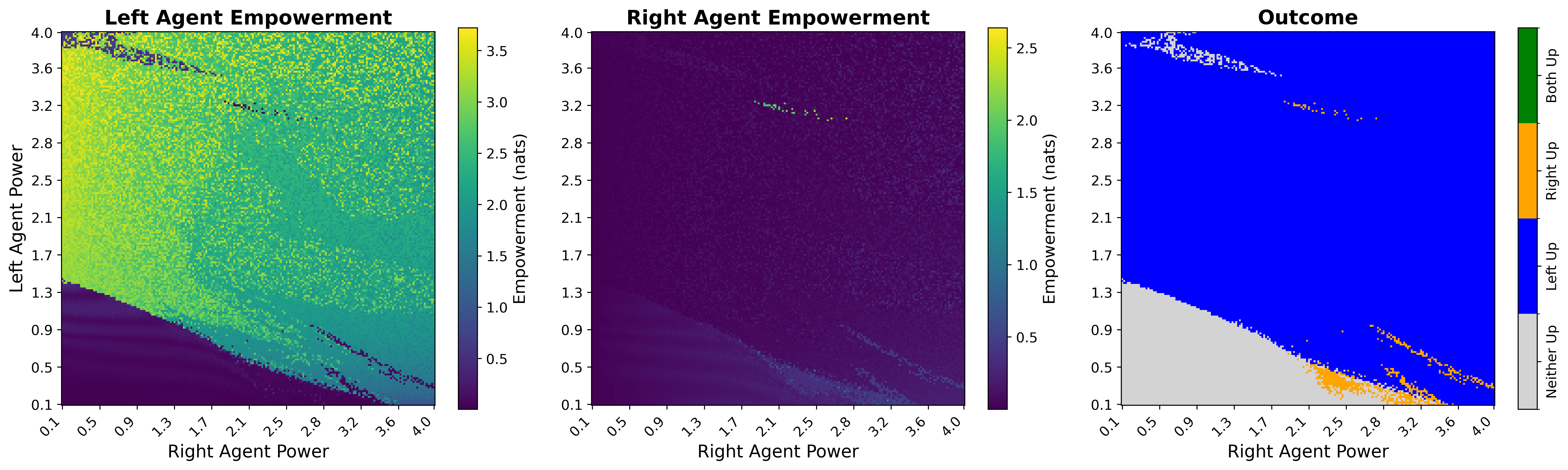}
        \caption{\textbf{Altruistic}: the right agent maximizes the empowerment of the left agent by following the policy \eqref{eq:altruistic-policy}. The left agent still follows the egoistic policy \eqref{eq:egoistic-policy}.}
        \label{fig:heatmap_altruistic}
    \end{subfigure}

    \caption{Heatmaps for the linked pendulum environment under two empowerment objectives (top: egoistic; bottom: altruistic; see subcaptions for policy definitions). The axes sweep the maximum torque available to each agent. Planning horizon is $1.3$~s with time step $\Delta t = 0.01$~s. In the outcome heatmaps, a pendulum is classified as ``up'' if its angle is within $\pm 1$ radian of vertical.}
    \label{fig:empowerment_heatmaps}
\end{figure}

Figure \ref{fig:empowerment_heatmaps} shows that the system exhibits distinct behavioral regimes as a function of the agents' relative torques. Under the egoistic policy (Figure \ref{fig:heatmap_egoistic}), when one agent is substantially stronger than the other, it tends to reach and hold the upright position while mechanically constraining the weaker agent through the tendon. This produces the asymmetric outcomes shown in Figures \ref{fig:left-up} and \ref{fig:right-up}. We did not observe either agent reaching upright under egoistic control with less than $\sim0.5$ torque.

When the agents have more comparable torque limits, a cooperative regime can emerge in which both pendulums reach upright simultaneously (Figure \ref{fig:both-up}). This regime yields high empowerment for both agents, although each agent's empowerment here is generally lower than in strongly asymmetric states where a single agent dominates (see the empowerment plots in Figure \ref{fig:heatmap_egoistic}).

We next consider a mixed-objective setting in which the left agent follows the egoistic policy while the right agent follows the altruistic policy targeting the left agent's empowerment (Figure \ref{fig:heatmap_altruistic}). In this case, the left agent reaches the upright state across a substantially larger region of the torque space, including regimes where it would not be able to self-right under egoistic control alone. In particular, increasing the right agent's available torque expands the set of low-power conditions under which the left agent can still attain high empowerment.

These experiments demonstrate that multi-agent empowerment is sensitive to both the relative control capabilities of agents and their policies. Switching from egoistic to altruistic maximization can qualitatively reshape the reachable outcome landscape, enabling weaker agents to reach high-empowerment states that are unreachable under purely egoistic control.

\subsection{Vicsek Flock}\label{sec:vicsek}

We next evaluate our method in a genuinely collective setting: a two-dimensional Vicsek-style flock with $N=125$ agents, in contrast to the two-agent pendulum system studied above. Following the implementation in \citet{Ferretti_2022}, each agent moves at constant speed with baseline Vicsek dynamics \citep{vicsek_1995} biasing its heading toward local neighbor alignment. On top of this alignment term, each agent controls its own angular acceleration; this added control is new in our model compared to standard Vicsek flocks. As in the pendulum setting, each agent's sensor $C^{(n)}$ selects its angle, so empowerment reflects each agent's capacity to control its own heading. Unlike the linked pendulums, interactions here are local, time-varying, and mediated through the group as a whole, so any large-scale structure must emerge from collective dynamics rather than direct pairwise coupling.

\begin{figure}
    \centering

    \begin{subfigure}[t]{0.24\textwidth}
        \centering
        \fbox{\includegraphics[width=\linewidth]{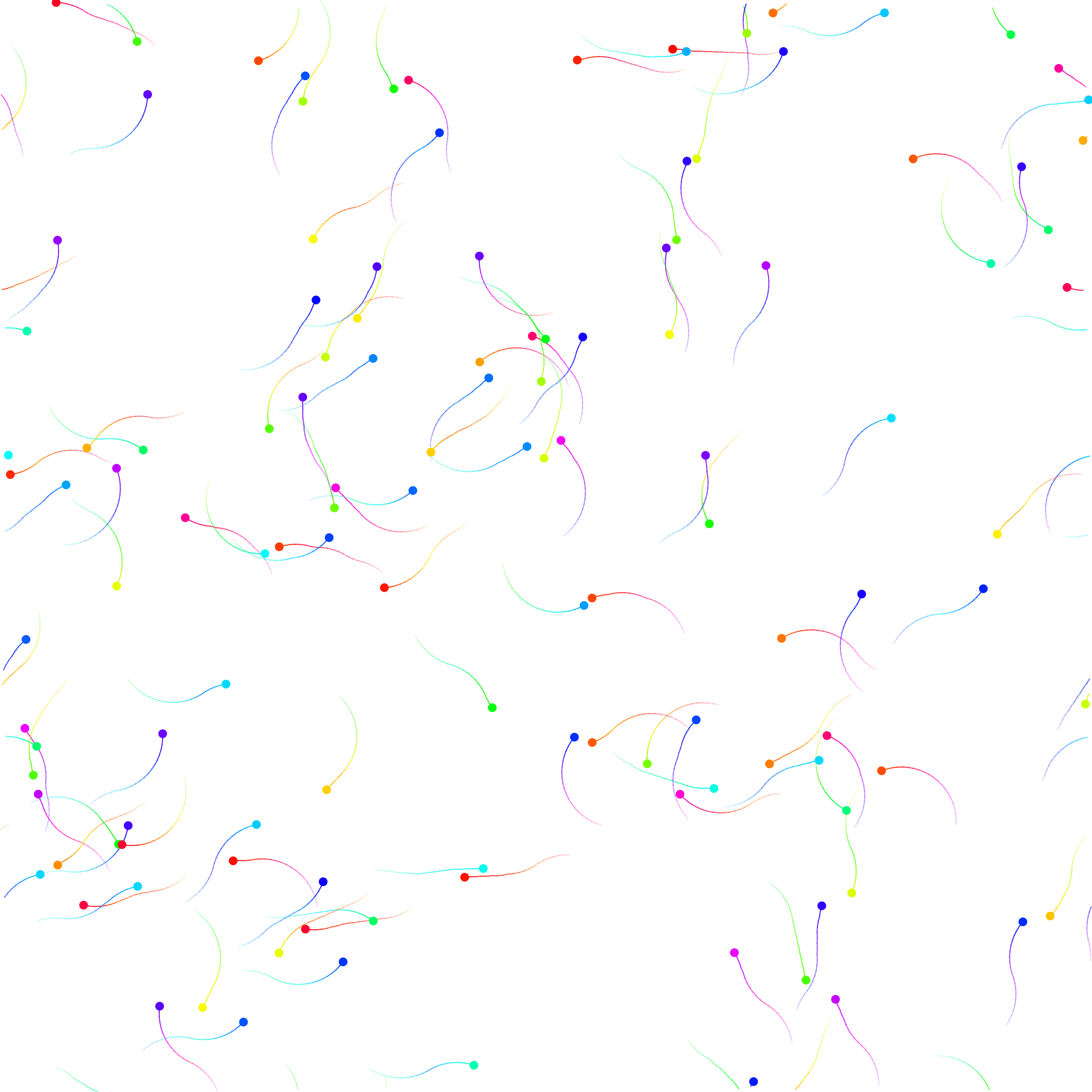}}

        \fbox{\includegraphics[width=\linewidth]{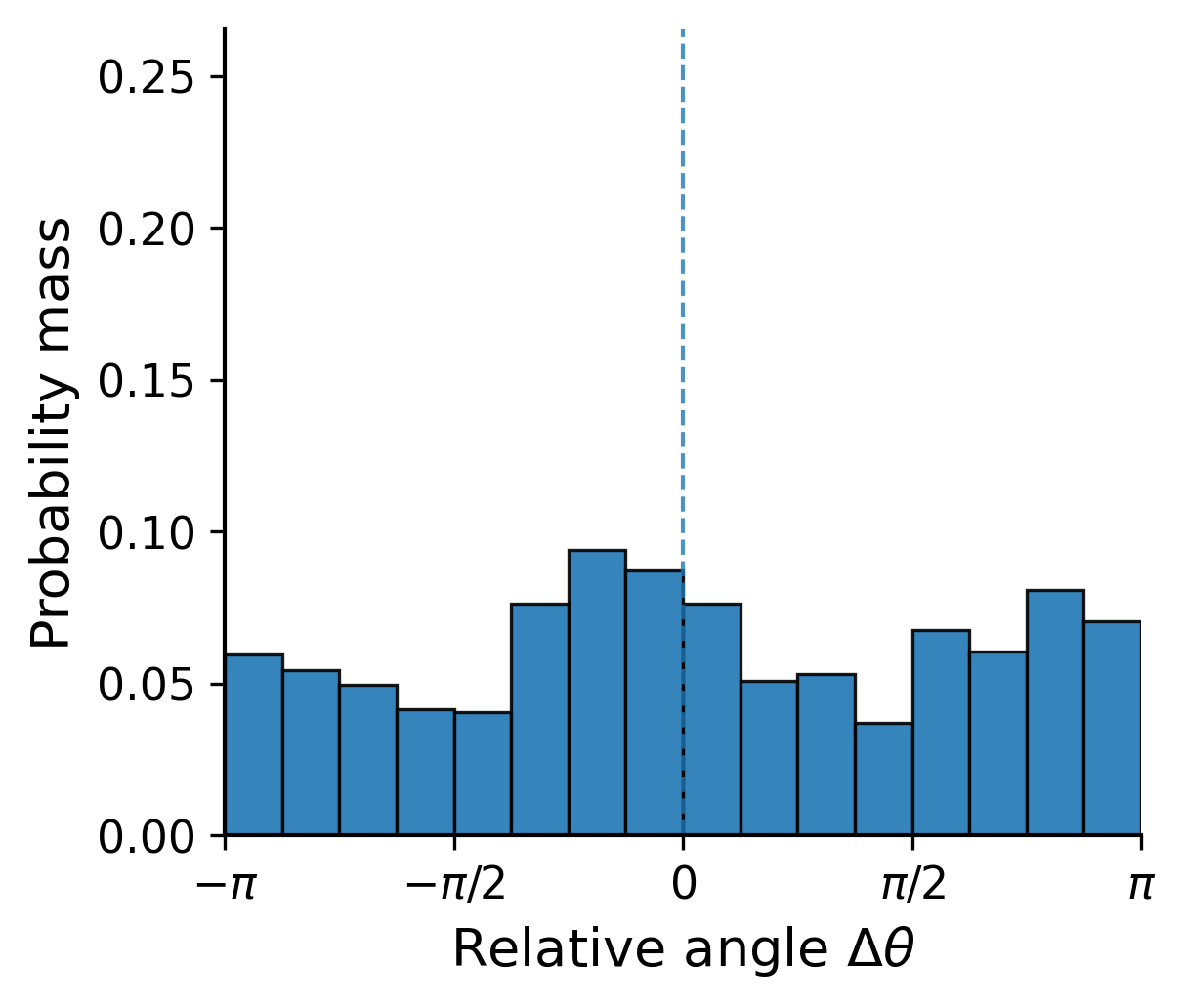}}
        \caption{$1$ (s)}
        \label{fig:plot1}
    \end{subfigure}
    \hfill
    \begin{subfigure}[t]{0.24\textwidth}
        \centering
        \fbox{\includegraphics[width=\linewidth]{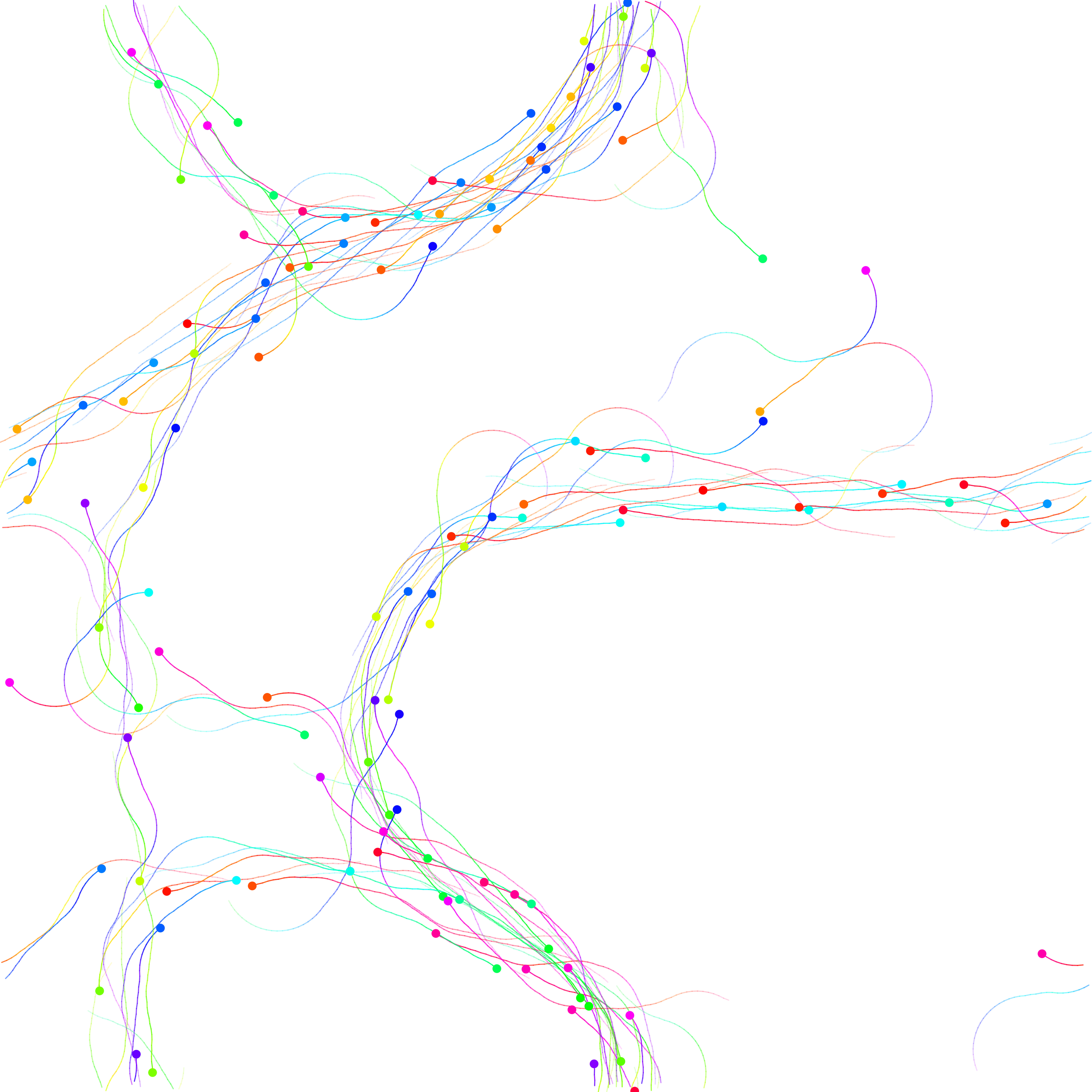}}

        \fbox{\includegraphics[width=\linewidth]{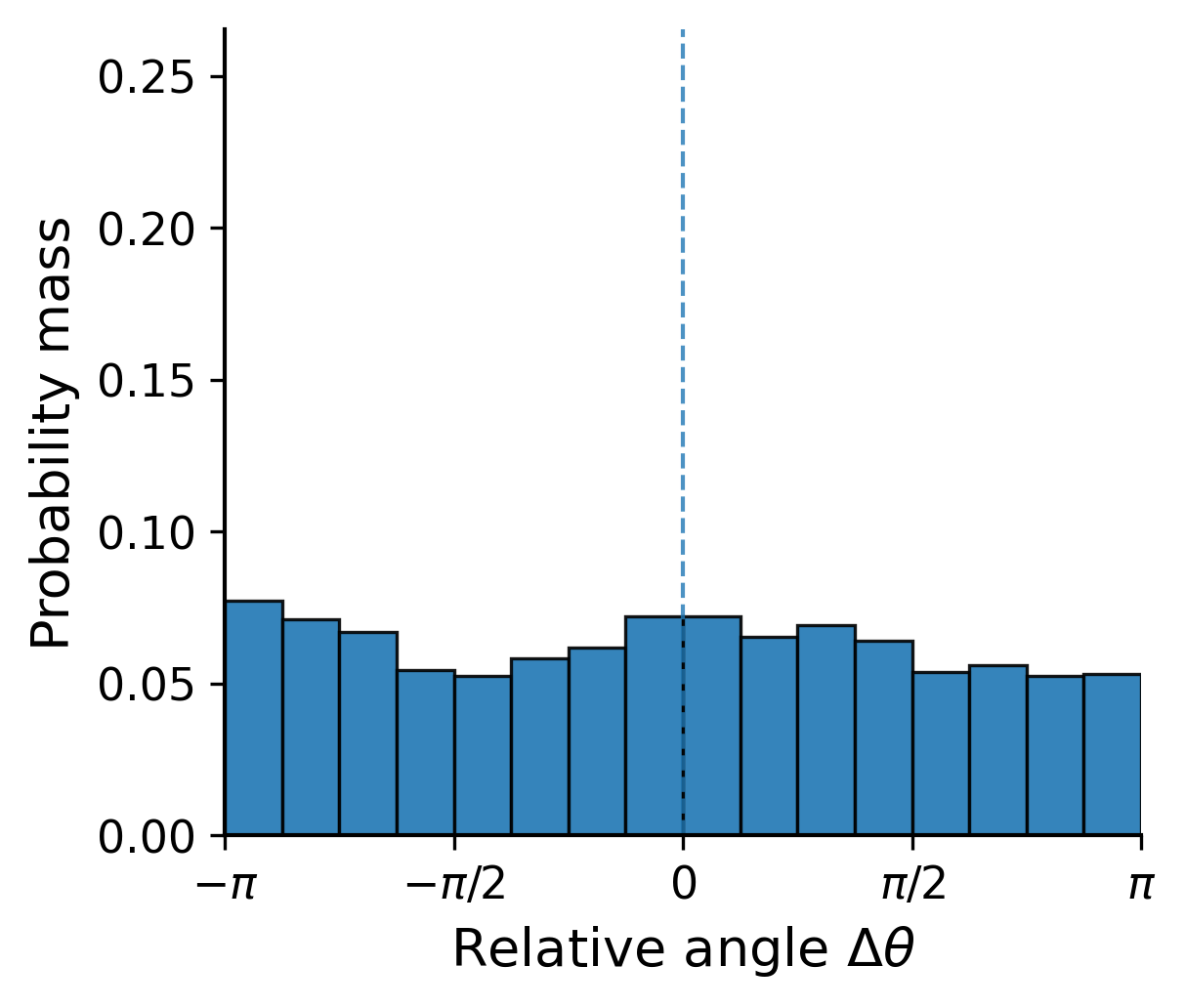}}
        \caption{$50$ (s)}
        \label{fig:plot2}
    \end{subfigure}
    \hfill
    \begin{subfigure}[t]{0.24\textwidth}
        \centering
        \fbox{\includegraphics[width=\linewidth]{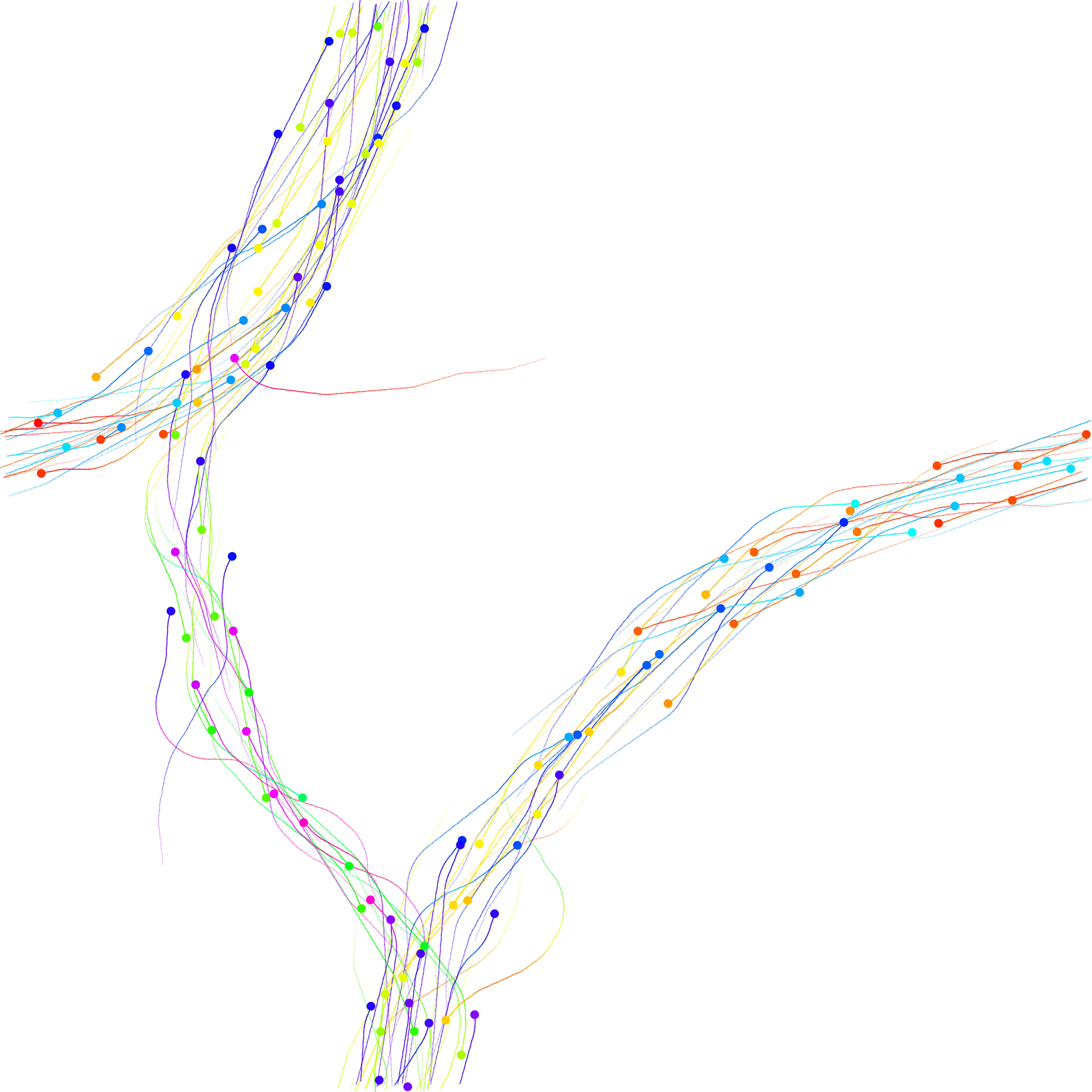}}

        \fbox{\includegraphics[width=\linewidth]{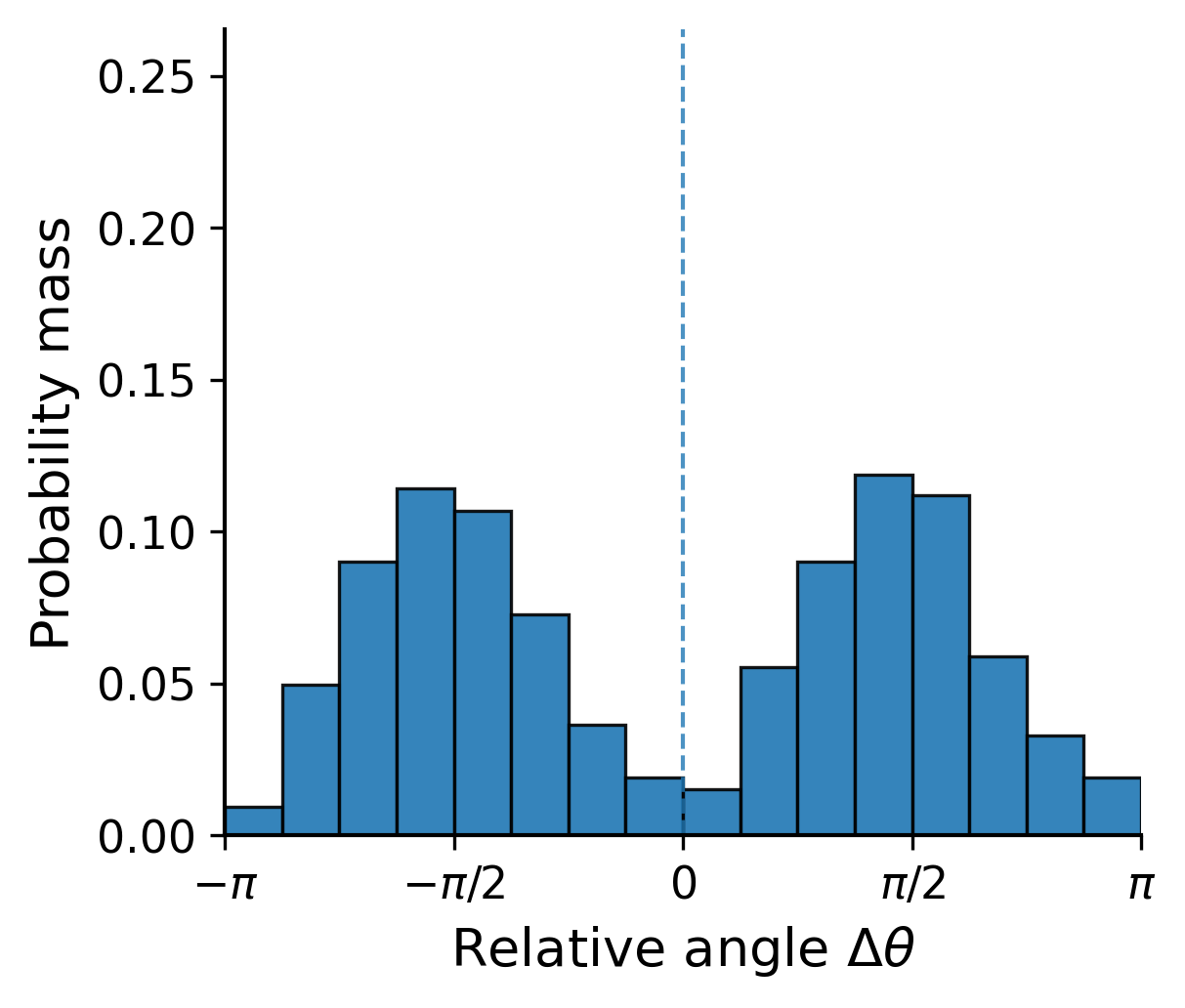}}
        \caption{$100$ (s)}
        \label{fig:plot3}
    \end{subfigure}
    \hfill
    \begin{subfigure}[t]{0.24\textwidth}
        \centering
        \fbox{\includegraphics[width=\linewidth]{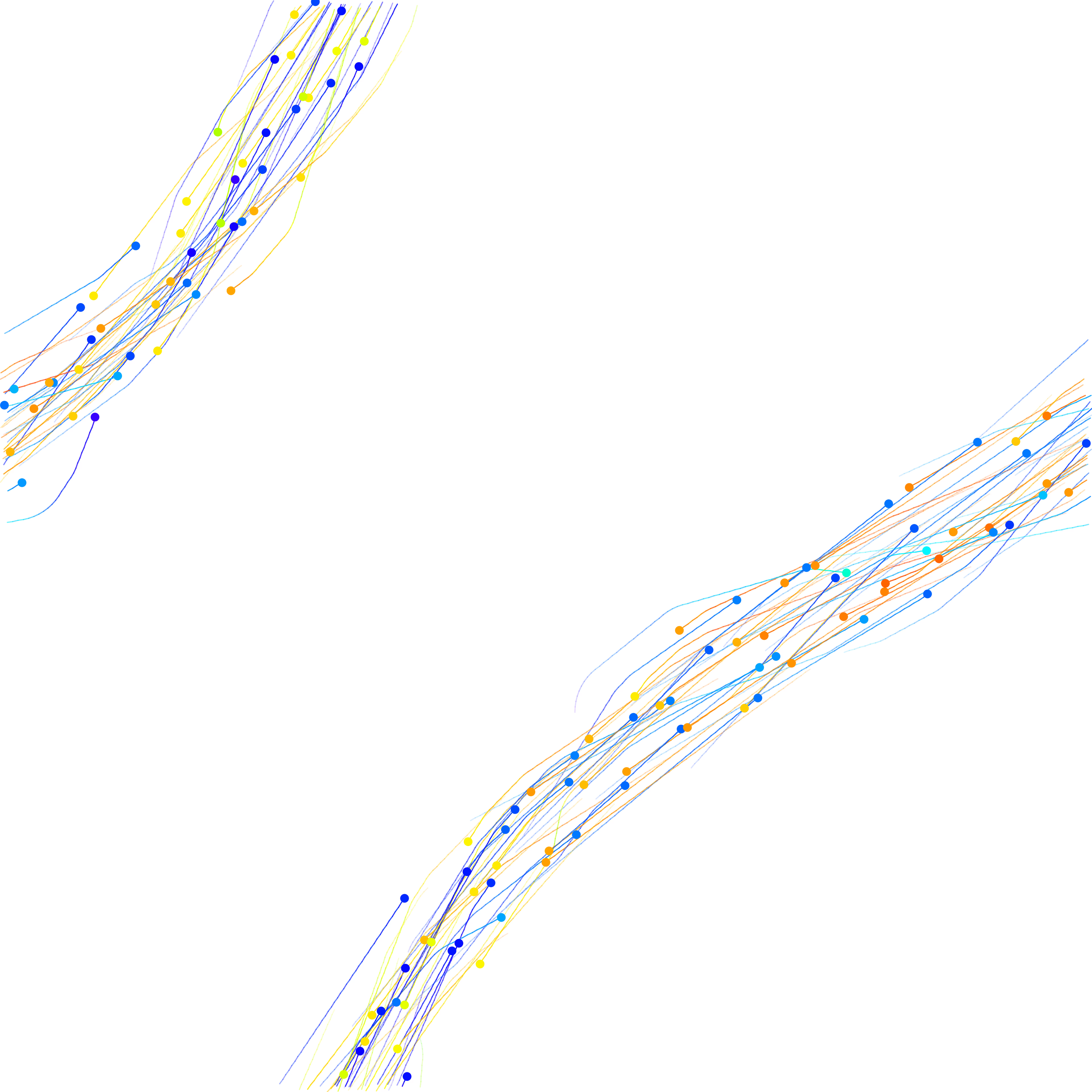}}

        \fbox{\includegraphics[width=\linewidth]{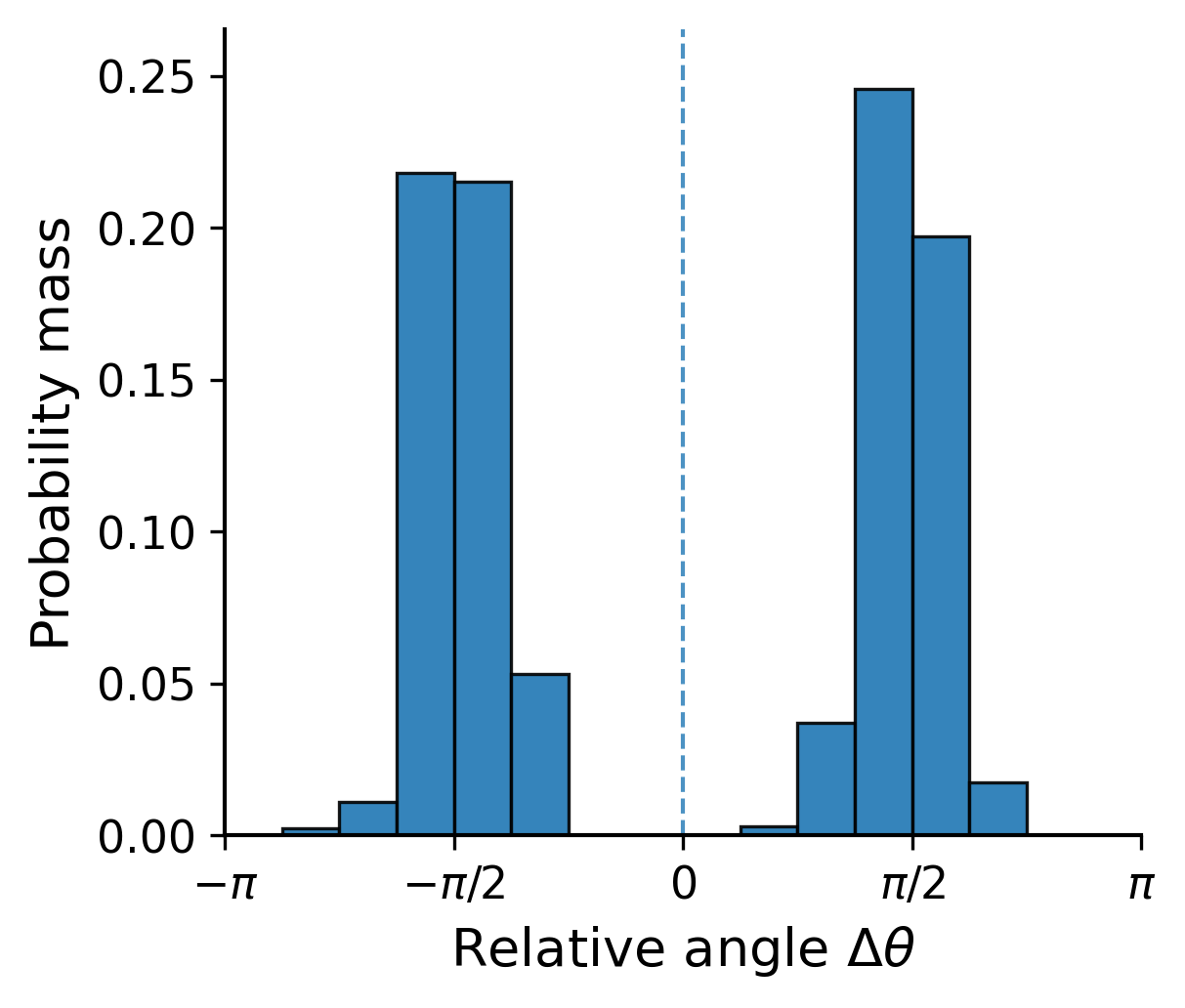}}
        \caption{$200$ (s)}
        \label{fig:plot4}
    \end{subfigure}

    \caption{
    Global structure emerges under egoistic empowerment maximization in a flock of $N=125$ agents. In each column, the top panel shows a snapshot of the flock at the indicated time: dots mark agent positions, trails show recent paths, and color encodes heading. The bottom panel shows the corresponding distribution of headings $\Delta \theta$ relative to the flock mean, evolving from approximately uniform at early times to an approximately bimodal distribution with two opposing preferred directions. Animations of the flock dynamics are available on the \href{https://fanciful-palmier-2afc73.netlify.app/}{project website}.
    }
    \label{fig:flock-global-structure}
\end{figure}

\begin{figure}[ht]
    \centering

    \begin{subfigure}{0.48\textwidth}
        \centering
        \includegraphics[width=\linewidth]{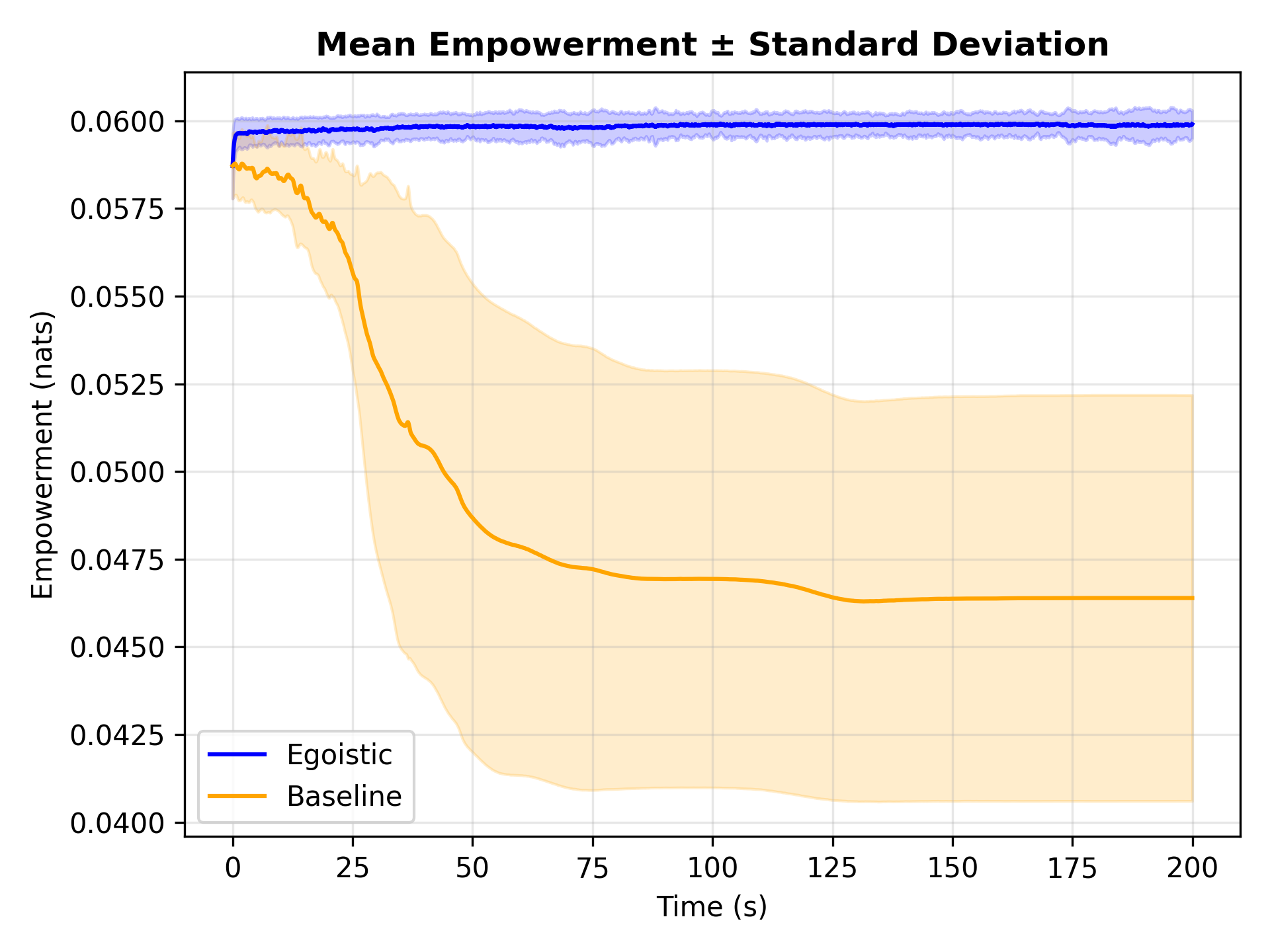}
        \caption{}
        \label{fig:flock-empowerment}
    \end{subfigure}
    \hfill
    \begin{subfigure}{0.48\textwidth}
        \centering
        \includegraphics[width=\linewidth]{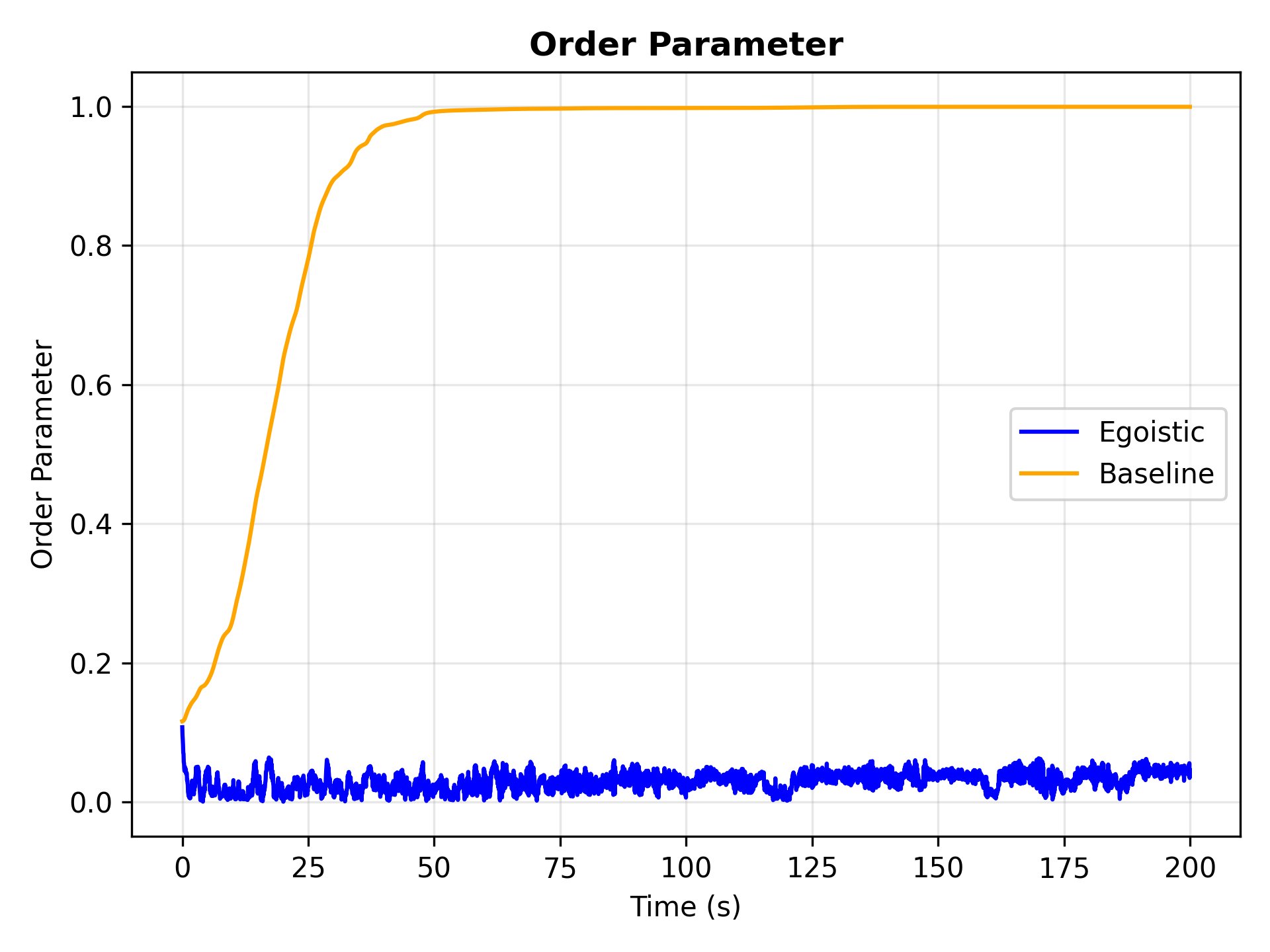}
        \caption{}
        \label{fig:flock-order}
    \end{subfigure}
    
    \caption{\ref{fig:flock-empowerment} Average flock empowerment over time under egoistic control and baseline Vicsek dynamics. \ref{fig:flock-order} Flock order parameter over time. Egoistic empowerment sustains high individual control and suppresses consensus alignment, while the baseline Vicsek dynamics drive the flock toward complete co-alignment.}
    \label{fig:flock_stats}
\end{figure}

Figure \ref{fig:flock_stats} compares the baseline Vicsek dynamics with egoistic empowerment-driven control. Under the baseline Vicsek dynamics, the average empowerment decreases steadily over time (Figure \ref{fig:flock-empowerment}) as the flock converges toward a highly aligned configuration. Agents effectively lose individual control as they become increasingly aligned with the collective. In contrast, when each agent acts to maximize its own empowerment, the average empowerment in the flock remains substantially higher throughout the simulation.

This difference is mirrored in the flock order parameter defined in \cite{vicsek_1995} (Figure \ref{fig:flock-order}). Baseline Vicsek dynamics rapidly drive the system toward global alignment, with the order parameter approaching $1.0$. By contrast, egoistic empowerment suppresses convergence to a single shared heading and keeps the order parameter near $0.0$. In other words, empowerment-seeking agents resist the loss of individual directional freedom that accompanies consensus.

Low order under the egoistic policy does not imply the absence of structure. Figure \ref{fig:flock-global-structure} shows that the flock self-organizes into coherent large-scale spatial patterns despite remaining globally disordered in heading. Starting from uniformly random headings, the population evolves through transient filamentary formations before settling into a single band configuration that wraps around the periodic domain. The order parameter is therefore spatially inhomogeneous, with local alignment likely approaching unity inside the band and dropping discontinuously at the boundaries between counter-propagating regions, even though its global average remains near zero.

The heading distributions in the lower row of Figure \ref{fig:flock-global-structure} clarify this effect. Rather than converging to a unimodal distribution centered on a common direction, the egoistic policy produces an approximately bimodal heading distribution with two opposing preferred directions. Agents maintain high empowerment by splitting into two counter-propagating streams rather than dispersing or co-aligning.

Together, these results show that egoistic empowerment maximization is sufficient to generate non-trivial large-scale structure in a multi-agent system, without any externally prescribed coordination.

\section{Discussion}\label{sec:discussion}

We formulated multi-agent empowerment as an interference channel, in which each agent's actions form a signal toward its own future state, and the other agents' actions appear as structured interference. The joint problem becomes a non-cooperative game, whose Nash equilibrium is computed by iterative water-filling, giving the first tractable, principled framework for empowerment in coupled, continuous-state multi-agent systems. In our formulation, the block-Jacobian $\mathbf{F}_t$ scales quadratically in the number of agents, but in spatially distributed systems, empirically, most off-diagonal coupling blocks are negligible (Appendix~\ref{sec:sparse-vicsek}). Thus the iterative water-filling computation reduces to each agent's local interaction neighborhood, which makes our approach tractable even for large collectives.

In two qualitatively different settings---two linked pendulums (Section~\ref{sec:pendulum}) and a 125-agent Vicsek-style flock (Section~\ref{sec:vicsek})---empowerment maximization alone produces structured group  behavior with no extrinsic coordination objective required. The character of that behavior depends on the coupling, on the egoistic versus altruistic choice, and on the sensors through which empowerment is measured. Varying parameters in the pendulums system yields dominance, cooperation, or altruistic support. In the flock, egoistic empowerment drives the population away from consensus alignment and into spatially organized counter-propagating bands. In this structured non-consensus state, local alignment spreads across the population without a global consensus. Altruism toward a single agent instead concentrates the group's directional freedom in that agent (Appendix~\ref{sec:altruistic-flock}). We expect different sensors and coupling topologies to produce qualitatively different but equally organized phases.

This framework brings together two fields that have arrived at related questions from opposite ends. Machine learning and robotics are entering an era of interacting autonomous agents---multi-agent reinforcement learners \citep{jaques2019socialinfluenceintrinsicmotivation}, populations of large language model agents \citep{guo2024large, tran2025multiagentcollaborationmechanismssurvey}, or multi-robot teams \citep{balch2002behavior, vorotnikov2018multi}---each pursuing its own local objective and producing collective outcomes that no one designed. The same problem appears in physics of living systems, where decades of active matter research \citep{ramaswamy2010,marchetti2013} have shown how self-propelled particles can produce flocks, bands, swirls, and jams from a handful of simple local rules. The crucial difference between these two fields is that active matter's rules are heuristic and purposeless, while autonomous agents' rules are functional and derived from an objective. Our framework unifies the two by giving each agent an information-theoretic function---increase its own future capacity to act---and computing the resulting joint dynamics tractably. Parameterized by the sensor $C^{(n)}$, the power budget $P^{(n)}$, the planning horizon, and the choice of a target agent, this function turns the collective into a tunable, functioning matter in the sense recently articulated by \citet{liu2025nitmb}: a continuous family of microscopic objectives mapping to a continuous family of emergent collective phases. By analogy with the active-matter literature, we expect this model class to support a broad and largely unmapped landscape of collective behaviors, of which the counter-propagating bands of Section~\ref{sec:vicsek} are one example. Our approach opens the door to charting this landscape systematically, which is likely to be a program for both fields for many years.

In addition to producing structure, empowerment maximization also {\em potentiates} the collective, driving it into configurations where each agent retains the capacity to act and where small directing signals can elicit behavior the system could not otherwise produce. A similar recently identified biological example is transport by ants, where cooperating ants self-organize into configurations from which a single transiently informed, ``empowered'' individual can redirect the entire group, enabling labyrinth navigation no individual ant could solve \citep{gelblum2015}. Our altruistic pendulum demonstrates a simpler version of a similar phenomenon: an agent's altruistic action moves the joint system into a configuration where the other agent reaches a state inaccessible under egoistic control alone. We hypothesize that this generalizes: high-empowerment configurations are uniquely ready to be steered toward external goals, making empowerment maximization a candidate preconditioning step for goal-directed group control.

We deliberately focused this paper on three contributions: (1) a formulation of multi-agent empowerment as an interference channel, (2) an efficient solution via IWF, and (3) a characterization of the behaviors that emerge purely from this objective. Combining multi-agent empowerment with extrinsic task rewards is a natural follow-up we leave to future work; establishing what behavior multi-agent empowerment in the continuum produces on its own is a prerequisite for understanding what it adds once paired with a task.

Overall, our work opens several future research directions. Most immediately, implementing end-to-end training with multi-agent empowerment as a differentiable objective would be the direct next step. Beyond that, mapping the phase diagram of intrinsically motivated matter---varying sensor, coupling topology, planning horizon, and population---would continue the work this paper has begun to sketch. Empirically, fitting such models to flock and swarm data would test whether intrinsic motivation explains observed collective phenomena better than alignment-only active matter models. Finally, operationalizing the potentiation hypothesis into a two-stage controller---empowerment preconditioning followed by goal-directed steering---would extend the framework from explanation to design in living and synthetic collectives.

\appendix

\section{Appendix}

\subsection{Sparse Interaction in Vicsek Flock}\label{sec:sparse-vicsek}

In the Vicsek flock environment the cross interaction between agents is related to the distance between them. We observe in Table~\ref{tab:force_distance} that agents which are far apart have negligible influence on one another compared to agents that are close together. This indicates that many off diagonal blocks in $\mathbf{F}_t$ could be ignored in IWF.
\begin{table}[h]
\centering
\caption{Mean norm of off diagonal blocks of $\mathbf{F}_t$ in \eqref{eq:block_sensitivity} in the Vicsek environment ($N=125$) aggregated over $3$ random states. Norms are binned by distance.}
\label{tab:force_distance}
\small
\begin{tabular}{lcc}
\toprule
Distance Bin & Mean $\|F_t^{(n,m)}\|$ & Count \\
\midrule
$[0.000,\ 1.760)$ & $1.5019 \times 10^{-4}$ & 4562 \\
$[1.760,\ 3.521)$ & $2.0586 \times 10^{-7}$ & 13816 \\
$[3.521,\ 5.281)$ & $1.8885 \times 10^{-12}$ & 21346 \\
$[5.281,\ 7.041)$ & $1.9751 \times 10^{-17}$ & 6774 \\
\bottomrule
\end{tabular}
\end{table}

\subsection{Altruistic Vicsek Flock}\label{sec:altruistic-flock}

\begin{figure}
    \centering
    \includegraphics[width=0.5\linewidth]{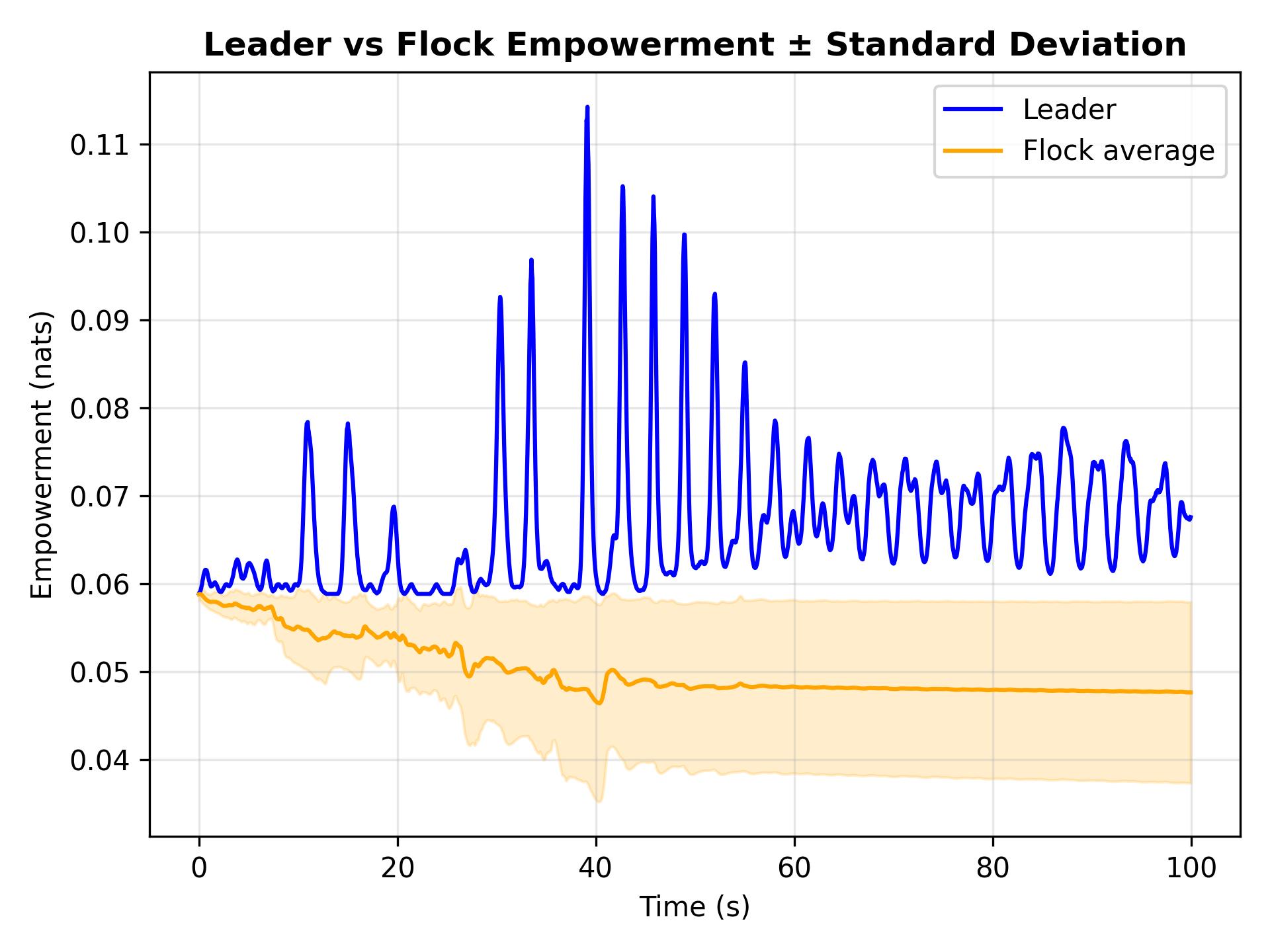}
    \caption{Altruistic scenario for the Vicsek Flock ($N=100$). One agent is selected as a ``leader'' who follows the egoistic policy (\eqref{eq:egoistic-policy}) while all other agents follow the altruistic policy (\eqref{eq:altruistic-policy}) to increase the empowerment of the leader.}
    \label{fig:altruistic-vicsek}
\end{figure}

When the collective ($N=100$) maximizes the empowerment of a single designated agent, that agent's empowerment indeed increases (Figure~\ref{fig:altruistic-vicsek}), and a behavior emerges: the ``leader'' agent moves in opposition to the rest of the flock, while the remaining agents align with each other. The collective sacrifices its own directional freedom to maximize the leader agent's ability to be different from the group. This is qualitatively distinct from the bimodal counter-propagating bands observed under purely egoistic empowerment (Figure~\ref{fig:flock-global-structure}). An animation of this scenario is available on the \href{https://fanciful-palmier-2afc73.netlify.app/}{project website}.

\section{Acknowledgements}

T.S. was supported by the Koh Family Scholarship. I.N. was supported, in part, by the Simons Foundation Investigator Award. S.T. was supported in part by NSF Award (2513350) and Alliance Innovation Lab in Silicon Valley.

\bibliography{collas2026_conference}
\bibliographystyle{collas2026_conference}

\end{document}